\documentclass[10pt,twocolumn,letterpaper]{article}

\usepackage{iccv}
\usepackage{times}
\usepackage{epsfig}
\usepackage{graphicx}
\usepackage{amsmath}
\usepackage{amssymb}

\usepackage[pagebackref=true,breaklinks=true,letterpaper=true,colorlinks,bookmarks=false]{hyperref}

\iccvfinalcopy 

\def\iccvPaperID{9217} 

\ificcvfinal\fi

\begin{document}

\title{ Glimpse-Attend-and-Explore: Self-Attention for Active Visual Exploration}

\author{Soroush Seifi\footnotemark[1] \hspace{0,5cm} Abhishek Jha\thanks{Equal contribution.}  \hspace{0,5cm}  Tinne Tuytelaars\\
PSI, ESAT, KU Leuven\\
Kasteelpark Arenberg 10, 3001 Leuven\\
{\tt\small firstname.lastname@esat.kuleuven.be}
}

\maketitle
\ificcvfinal\thispagestyle{empty}\fi
\begin{abstract}

Active visual exploration aims to assist an agent with a limited field of view to understand its environment based on partial observations made by  choosing the best viewing directions in the scene. Recent methods have tried to address this problem either by using reinforcement learning, which is difficult to train, or by uncertainty maps,
which are task-specific and can only be implemented for dense prediction tasks. In this paper, we propose the Glimpse-Attend-and-Explore model which: (a)  employs self-attention to guide the visual exploration instead of task-specific uncertainty maps; (b) can be used for both dense and sparse prediction tasks; and (c) uses a contrastive stream to further improve the representations learned. Unlike previous works, we show the application of our model on multiple tasks like reconstruction, segmentation and classification. Our model provides encouraging results 
while being less dependent on dataset bias in driving the exploration. We further perform an ablation study to investigate the features and attention learned by our model. Finally, we show that our self-attention module learns to attend different regions of the scene by minimizing the loss on the downstream task. Code:
 \url{https://github.com/soroushseifi/glimpse-attend-explore}.

\vspace*{-0.5cm}
\end{abstract}

\begin{figure*}
  \begin{center}
      \includegraphics[width=\linewidth]{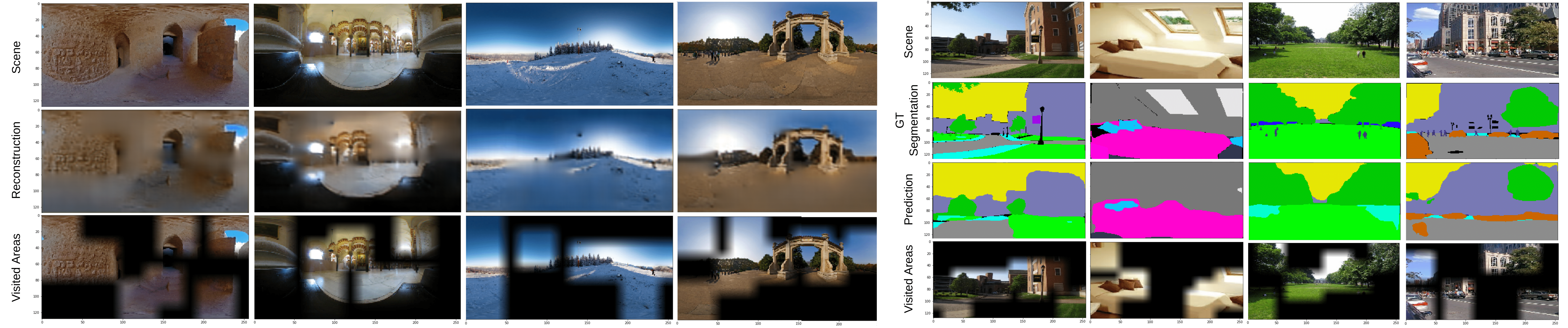}
   \end{center}
      \vspace*{-0.4cm}
      \caption{\textbf{Results generate by Glimpse Attend and Explore:} (Left) shows the scene reconstruction results on SUN360 dataset \cite{song2015sun}. (Right) shows the semantic segmentation results on ADE20k \cite{zhou2017scene}. Results are computed after taking 8 retina-like glimpses (bottom-row) containing 18\% of the pixels in the image.}
      \label{3streams}
      \vspace*{-0.25cm}
   \end{figure*}

\vspace*{-0.15cm}
\section{Introduction}
\vspace*{-0.15cm}
Most computer vision methods rely on datasets captured by human photographers \cite{lin2014microsoft, russakovsky2015imagenet}. Such data is biased towards the salient information showing up in predictable areas of the image (e.g.~image center). Besides, most computer vision methods assume full observability of the input image \cite{kendall2015posenet, girshick2014rich}. However, in a dynamic environment, an agent with a limited field of view/resource cannot fully observe its immediate 360\textdegree scene. This might cause the agent's camera to capture parts of the environment that are divergent to those seen in standard computer vision dataset, thus degrading agent's performance on different tasks. 

In this paper, we propose an active vision \cite{aloimonos1988active} method to autonomously explore and reason about a scene by sequentially gathering partial observations from it. Our method can be deployed in scenarios where an agent cannot view and process the whole scene due to limitations such as the agent's small field of view or limited transfer bandwidth between the camera and the processing unit. We simulate this by restricting our method to see small crops (called \textit{glimpses}) from the images in common computer vision datasets. Besides, we restrict the total number of glimpses the agent can see from a single image. At each time step, the agent has the freedom to change its viewing direction and take a new glimpse of the scene. Therefore, it is important that the agent selects the areas of the environment with the highest information gain for a given task. Given a set of training examples and an initial random glimpse for each example, our model learns a policy to select the next glimpses, hallucinates the unseen areas, and solves a task given the structural cues coming from the visited areas. 

While previous methods rely on reinforcement learning \cite{ramakrishnan2018sidekick,jayaraman2018learning}, reconstruction loss \cite{seifi2019look} and uncertainty measures \cite{seifi2020attend}, we employ the heatmaps generated by self-attention layers to guide the exploration. Contrary to previous works, we propose a unified two-stream architecture for different tasks such as image reconstruction, classification, and semantic segmentation and evaluate our method using several baselines. In addition, we show that the agent can build a richer representation of the environment by using contrastive learning. In this case, only during training, we use a pretrained encoder to produce the features for the full environment. Next, taking inspiration from \cite{chen2020exploring}, we train one of our network's streams to predict the full environment's features given only the visited glimpses. Finally, we perform an ablation study on the number of glimpses and our network architecture. Our contributions are as follows:
\begin{itemize}
\itemsep0em 
    \item We introduce a new self-supervised attention mechanism for active visual exploration.
    \item We propose a unified architecture for both sparse and dense prediction tasks.
    \item We show that our proposed attention mechanism disentangles location prediction from the auxiliary dense prediction task used in previous work \cite{seifi2020attend}. Consequently, a downstream task such as classification/regression may directly lead the attention mechanism. This makes the architecture two times faster and uses less than a tenth of the gpu memory compared to previous work. 
    \item We employ contrastive learning to train the network to reason beyond seen areas and build an even richer representation of the environment.
\end{itemize}

\begin{figure}
  \begin{center}
      \includegraphics[width=\linewidth]{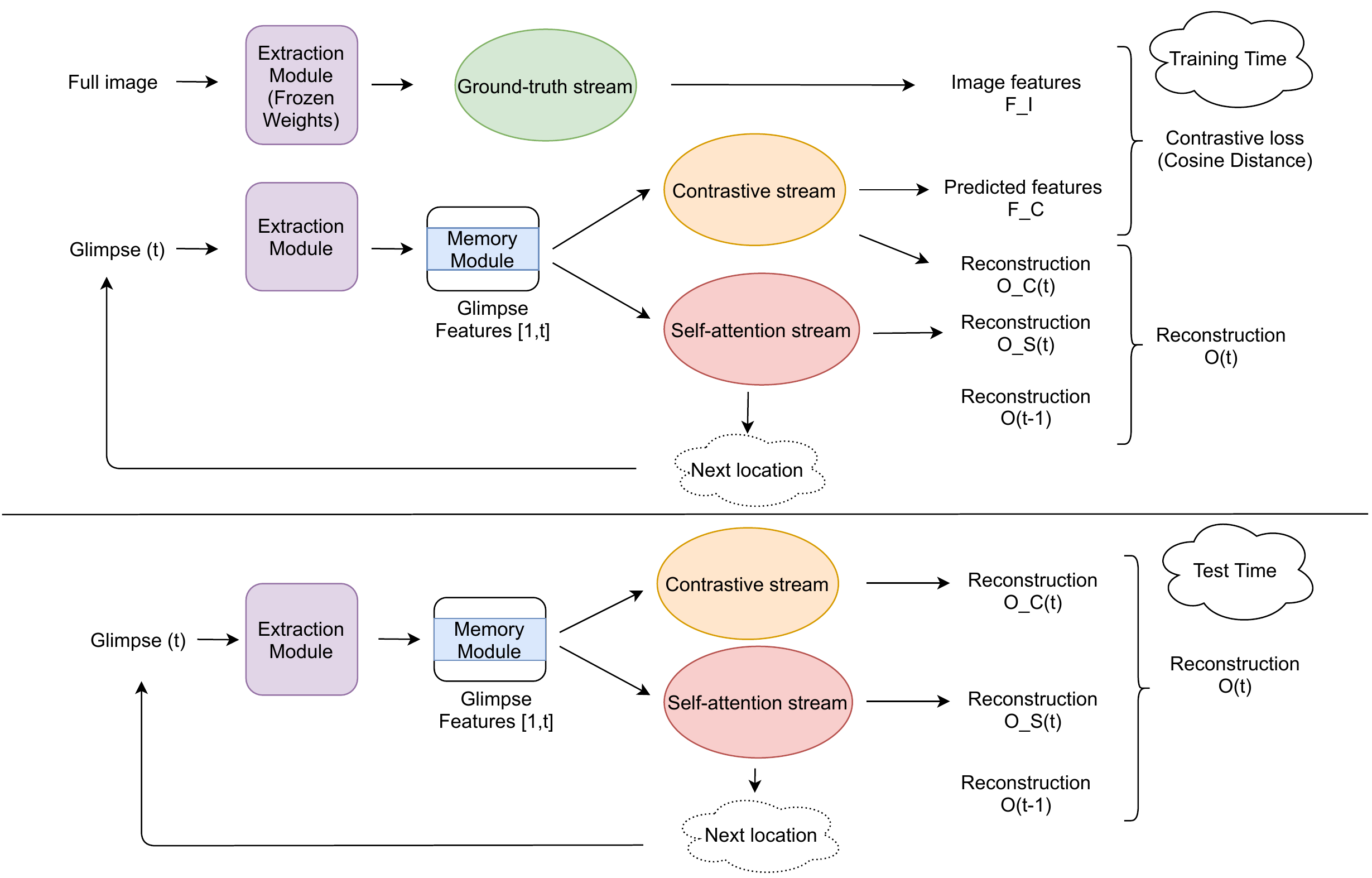}
   \end{center}
   \vspace*{-0.4cm}
      \caption{Architecture Overview.}
      \label{3streams}
   \end{figure}

\vspace*{-0.35cm}
\section{Related Work}
\vspace*{-0.15cm}
\textbf{Active Vision:} An agent with active vision can control its viewpoint to make a series of observations to subsequently improve its internal representation of the environment. Some of the earliest work \cite{aloimonos1988active} provides a general framework for this problem in low resource vision systems and camera control \cite{bajcsy1988active}. Recent work in this domain aims at learning view selection strategies to solve diverse tasks including object recognition \cite{mnih2014recurrent, ba2014multiple}, segmentation \cite{chai2019patchwork, long2015fully}, visual navigation \cite{song2018im2pano3d, chen2019learning, zhu2017target}, and pose estimation \cite{gartner2020deep, sock2019active}. Similar work to ours falls in the domain of active image understanding, in the subsequent subsection we provide a brief literature survey of the constituent modules of our model.

\textbf{Image Reconstruction:} Conditional image reconstruction based on partially observed image can be done either as an inpainting \cite{pathak2016context, yu2018generative} or outpainting task \cite{jayaraman2017unsupervised, ramakrishnan2018sidekick, seifi2019look}. While the amount of context available for inpainting is usually high, outpainting receives a partial view of the image as context. Jayaraman \etal \cite{jayaraman2017unsupervised} propose a view grid reconstruction as a pretext task to learn the 3D visual representation of a 2D view of the object. Ramakrishnan \etal \cite{ramakrishnan2018sidekick} introduces an action policy learning strategy to select a sequence of view grids to reconstruct the whole scene. Our work is in the same line as \cite{ramakrishnan2018sidekick}, where we sequentially select glimpses to reconstruct the whole scene, with the difference in the way we learn to select views. Similar to our work, \cite{ramakrishnan2018sidekick} employs the full image to help the training. However, different views of the view-grids in \cite{ramakrishnan2018sidekick} may have different FOV depending on their location in the gird. In this work, we fix the size of the glimpses and consequently scene coverage to always be the same. Besides, \cite{ramakrishnan2018sidekick} reduces the search space for the reinforcement learning training scheme by restricting the agent to always select from the neighbouring glimpses while our agent can change its viewing direction to anywhere in the scene. The closest to our work is \cite{seifi2019look}, which performs an attention-based view selection. The attention policy proposed by \cite{seifi2019look} learns to predict the image region with the highest reconstruction loss and thereby requires the loss value to be trained. The self-attention module of our proposed model uses gradients received from the next layer to train. Therefore, each layer of self-attention learns to attend the image regions specific to the downstream task.

\begin{figure*}
    \begin{center}
      \includegraphics[width=1.0\linewidth]{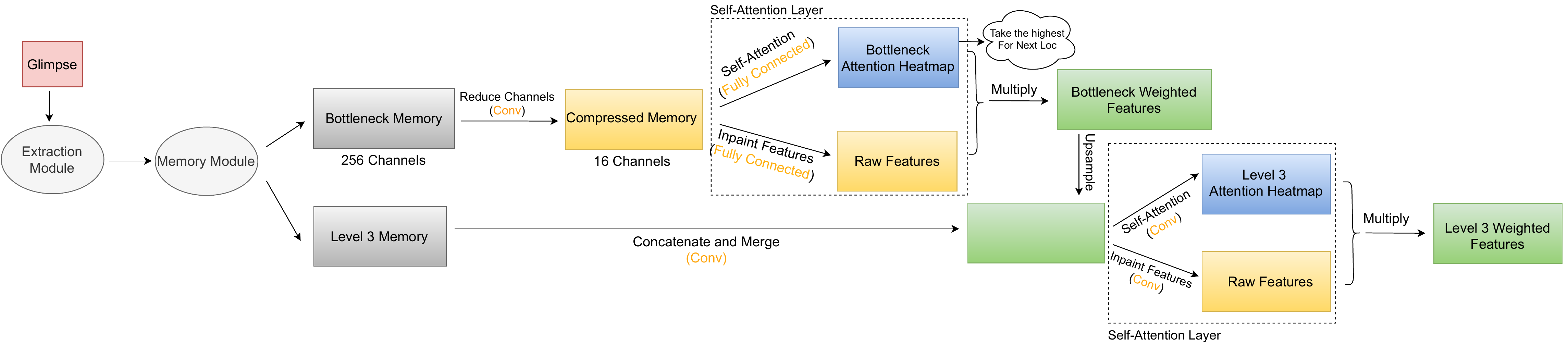}
    \end{center}
   \vspace{-0.5cm}
      \caption{First two levels of the self-attention stream's decoder.}
      \vspace{-0.3cm}
      \label{selfatt}
   \end{figure*}


\textbf{Semantic segmentation:} Conventional segmentation methods like FCN \cite{long2015fully}, U-net \cite{ronneberger2015u}, Segnet \cite{badrinarayanan2017segnet}  and Deeplab \cite{chen2017deeplab} have been successful for segmenting natural scenes and biomedical imaging, however, they cannot be used in environments with limited FOV where full observation of the environment is impossible.

Recent works aim to actively sample parts of the scene to provide segmentation masks, like \cite{han2018reinforcement}, which iteratively predicts an object and context boxes pairs to predict the segmentation mask around the object. However, it requires the initial location of the target object as an input. Chai \etal \cite{chai2019patchwork} uses an attention mechanism to guide the view selection policy to segment an object in the video stream. While we also use attention to guide the glimpse selection, we show the segmentation of multiple objects in the scene. We also evaluated our model on a more diverse set of datasets than these two previous works.

The most similar work to ours is \cite{seifi2020attend} which 
trains an attention mechanism by weighting the segmentation loss with an uncertainty map derived from the internal state of the network. 
Our method takes inspiration of using uncertainty maps, however, our attention mechanism is trained implicitly and does not directly rely on predicting the task-specific loss. The weights of this attention module are rather trained by the gradients coming from the successive layers. This allows the self-attention module to learn a different policy for each specific downstream task, without requiring an architectural change in the module itself, thereby making the module task-agnostic.
\vspace*{-0.13cm}

\textbf{Image classification:} Pioneering work in classification using active vision recurrent attention model (RAM) \cite{mnih2014recurrent} shows classification on cluttered MNIST dataset by learning a view selection policy. DRAM \cite{ba2014multiple} learns a deep RAM to show detection and classification of multi-digit MNIST dataset. Unlike these methods, we show classification on a challenging natural scene dataset. Our model also uses a spatial memory bank similar to \cite{seifi2019look, chai2019patchwork} to maintain a more expressive representation of the scene than the latent representation of a recurrent neural network \cite{mnih2014recurrent, ba2014multiple}, yet being more memory intensive.
\vspace*{-0.13cm}

\textbf{Attention mechanism:} Learning attention and saliency by optimization for specific tasks has shown significant improvements for both vision \cite{xu2015show, dosovitskiy2020image} and language tasks \cite{vaswani2017attention, devlin2018bert} over the non-attention counterparts. Our proposed self-attention module resembles the multiplicative attention proposed in Transformer model \cite{vaswani2017attention} and ViT \cite{dosovitskiy2020image}. However, for each layer, our self-attention module does not rely on an explicit query, but rather the attention is directly computed by processing the input features to that layer.
\vspace*{-0.13cm}

\textbf{Contrastive learning:} Hadsell \etal \cite{hadsell2006dimensionality} learns a representation by minimizing the distance between the positive pairs and maximizing it between negative pairs. Recently, a whole domain of self-supervised representation learning \cite{bachman2019learning, he2020momentum, chen2020simple, grill2020bootstrap, chen2020exploring} has adapted this contrastive learning formulation, by considering each instance as a separate class, to learn image features in the absence of ground truth labels.
In this paper, we use a similar loss formulation as \cite{grill2020bootstrap, chen2020exploring}, to minimize the distance between positive pairs in the representation space. As we will see in the experiment section, this specifically improves reconstruction quality.

\vspace*{-0.15cm}
\section{Method}
\vspace*{-0.15cm}
Our architecture consists of four main components. The `Extraction Module' encodes the features for each attended glimpse and the full image. The `Memory Module' gathers the features for all visited glimpses in spatial memory maps. The `Self-Attention Stream' employs self-attention layers to guide the exploration and to reason about the scene. The `Contrastive Stream' predicts the features for the full image given the partial observations. The task at hand (image reconstruction, semantic segmentation etc) is solved based on the outputs of the self-attention and contrastive streams. Finally, the locations to attend at each step is determined by the inner state of the self-attention module in that step. Figure \ref{3streams} provides an overview of our method.

\vspace*{-0.15cm}
\subsection{Extraction Module}
\vspace*{-0.15cm}
Following the architecture proposed in \cite{seifi2020attend} the extraction module receives a ``Retina-like glimpse". Such glimpses help saving the pixel-budget and processing power by scaling down on the areas that are located further from their center.

We use the first four layers of a (pretrained) Resnet-18 \cite{he2016deep} network to encode each retina-glimpse. Besides, only during training time, we use the same encoder to extract the full image features $F_{I}$ (with $I$ denoting the input image). Note that the network's weights are frozen when extracting features for the full image (Figure \ref{3streams}) and the gradients flow through this module only with glimpses.

\vspace*{-0.15cm}
\subsection{Memory Module}
\vspace*{-0.15cm}
Inspired by \cite{seifi2020attend} we employ spatial memory maps to keep the extracted features for the  glimpses visited up to the current time step. In particular, our memory module maintains 4 different matrices, one for each encoder level in the extraction module. We denote these matrices as `Level 1', `Level 2', `Level 3' (intermediate memories) and `Bottleneck' memory. Assuming that the full scene is an image of size $N \times M \times 3$, Level 1 memory would be a matrix of size $\frac{N}{2} \times \frac{M}{2} \times 64$ and Bottleneck memory would have the size $\frac{N}{16} \times \frac{M}{16} \times 256$. (Level 2: $\frac{N}{4} \times \frac{M}{4} \times 64$, Level 3: $\frac{N}{8} \times \frac{M}{8} \times 128$). 

After visiting a glimpse in the scene, the extracted features are stored in the corresponding location in these matrices. In case of overlap between two glimpses, these memories are updated with the features of the newest glimpse in the overlapping area. Note that if the agent visits all possible non-overlapping glimpses in the image, these matrices would contain the extracted features for the whole scene. However, since the number of glimpses is limited, these matrices in practice remain partially empty.

The contrastive and self-attention streams work on top of these partially filled matrices to fill-in the unvisited areas and build a representation for the whole scene.

\subsection{Contrastive Stream}
\vspace*{-0.15cm}
The purpose for this module is to predict full image features $F_{I}$ given only partial observations of the image $I$. Note that $F_I$ consists of 4 feature matrices corresponding to 4 encoder levels. We define a decoder symmetrical to the extraction module to create a U-net shaped architecture for the contrastive stream. While the extraction module encodes the features for each glimpse separately from the others, the contrastive stream operates on top of the `bottleneck memory' which contain the features for all glimpses visited until the current timestep. The features in the `intermediate' memories are used as skip connections to ease the gradient flow of the encoder/decoder architecture (See supplementary material for the detailed architecture).


We denote the predicted features after visiting all glimpses as $F_{C}$ and its corresponding loss by $L_{F_c}$. $L_{F_c}$ is calculated as the negative cosine distance between $F_{I}$ and $F_{C}$, equation \ref{eqn:contrastive_loss}. Note that $L_{F_c}$ summarises four loss terms for each encoder/decoder level. Minimizing this loss would train the network to assign similar representation to scenes with similar structure. 

Besides, we denote the output of this module at each step as $O_{C(t)}$ and its corresponding loss as $L_{C(t)}$. Depending on the task at hand $L_{C(t)}$ could be a L2 loss for reconstruction or cross-entropy loss for segmentation task.

\begin{equation}
\label{eqn:contrastive_loss}
L_{F_c} = - \frac{F_{C}}{\|F_{C}\|} \cdot \frac{stopgrad(F_{I})}{\|stopgrad(F_{I})\|}
\end{equation}

Note that our method is different from the contrastive learning framework introduced in \cite{chen2020simple} as we do not provide negative examples for the contrastive stream. We follow the method in \cite{chen2020exploring} where stopping the gradient on target image's features ($F_I$) eliminates the need for large batches and negative examples. As we will see in the experiments part, particularly in our settings where the goal is to hallucinate the unseen parts of the scene, it is better not to provide negative examples for the contrastive stream. This is due to the fact that two semantically different scenes might share features for large parts of their environment such as sky/walls for indoor/outdoor scenes. Therefore, pulling the features away for those scenes may not be suitable for
learning the representations for a task such as reconstruction.

\begin{figure}[h!]
  \begin{center}
      \includegraphics[width=\linewidth]{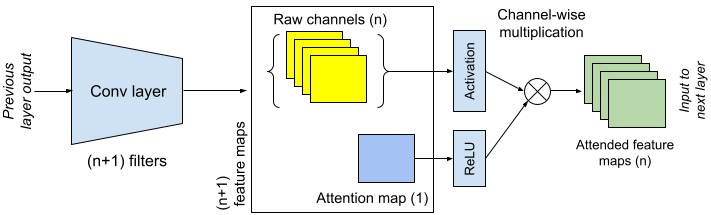}
   \end{center}
   \vspace*{-0.3cm}
      \caption{\textbf{Self-attention module:} Using an additional filter in the convolution layer, we predict an extra channel (attention map) along with the feature map (raw features). We apply ReLU activation \cite{nair2010rectified} on this attention map to make it non-negative. This attention map is then used as weight and multiplied to the raw feature maps to get attended feature maps. These attended feature maps are then passed to the next layer as input. }
      \label{fig:attention_module_image}
      \vspace*{-0.5cm}
   \end{figure}

\begin{table*}

\begin{center}
\begin{tabular}{|l|p{1.2cm}|p{1.2cm}|p{2cm}|p{2cm}|p{2cm}||p{2cm}|p{2cm}|}
\hline
Method/Dataset& Self-attention & Contrast & SUN360 \cite{song2015sun} (RMSE) $\downarrow$  & MS-COCO \cite{lin2014microsoft} (RMSE) $\downarrow$ & ADE20k \cite{zhou2017scene} (mPA \%) $\uparrow$ & COCO-Stuff \cite{caesar2018coco} (mPA \%) $\uparrow$ \\
\hline\hline
Ours & \checkmark & \checkmark & \textbf{33.8} & \textbf{40.3} & \textbf{52.4} & \textbf{47.8}  \\
\hline
No Contrast & \checkmark & x& 34.2 & 43.7 & 52.3 & 47.5 \\
\hline
Random glimpse (SA) & \checkmark & \checkmark & 39.4 & 47.8 & 51.5 & 46.0 \\
\hline
Random glimpse (No SA) & x & \checkmark & 38.6 & 48.5 & \textbf{52.4} & 47.7 \\
\hline
Random glimpse (No SAC) & x & x & 39.2 & 48.4 & 51.7 & 46.4 \\
\hline
\end{tabular}
\vspace{0.15cm}
\caption {\textbf{Evaluation of modules:} Performance comparison of different variants of our model by ablating individual modules, where  SA is Self-Attention stream, and SAC: Self-Attention stream + Contrast stream)}
\label{tab:baseline}
\end{center}
\vspace*{-0.7cm}
\end{table*}
\vspace*{-0.15cm}
\subsection{Self-Attention Stream}
\vspace*{-0.15cm}
The self-attention stream, figure \ref{selfatt}, has a similar architecture to the contrastive stream. However, at each decoder level it predicts an extra uncertainty channel (attention heatmap). This heatmap is then multiplied by the predicted features of that decoder level (figure \ref{fig:attention_module_image}). This way, while decoding, specific locations of the scene get a higher weight and thus a higher importance for solving the final task. Therefore, the heatmaps generated by the self-attention module are good indications of which regions are the most important ones to attend.

In our experiments, we use the bottleneck attention heatmap (figure \ref{selfatt}) to select the location for the next glimpse. This heatmap is generated using a fully-connected layer which takes into account all the activations in the bottleneck feature memory encoding the highest abstraction of the input scene. Besides, each pixel in this heatmap represents the importance of a $16 \times 16$ area in the scene. Consequently, we take the next glimpse from a previously unvisited area which has the highest activation in this attention map. 

Figure \ref{selfatt} illustrates the first two levels of the self-attention stream's decoder. Upsampling to the higher levels is done in a similar way as depicted for these two levels. As mentioned earlier, the contrastive stream follows a similar architecture but without the attention heatmaps. We denote the output of this module as $O_{S(t)}$ and its loss by $L_{S(t)}$.
\vspace*{-0.15cm}
\subsection{Final task and the Network's Architecture}
\vspace*{-0.15cm}
Depending on the nature of the final task at hand the decoder part of the network can be designed differently to work more efficiently. For a dense prediction tasks such as image reconstruction and semantic segmentation, we use the full decoder described for the contrastive and self-attention streams.

However, for a classification/regression task, unlike previous methods \cite{seifi2019look,seifi2020attend}, training of the attention mechanism is not dependant on a dense prediction task's loss and relies only on the bottleneck attention heatmap (figure \ref{selfatt}).

In this case, the self-attention and contrastive streams are only employed at the level of bottleneck features and the rest of the decoder for both modules as well as all the intermediate memories are removed from the pipeline. As we will see in the experiments section, this makes the network to work faster and use less GPU memory.

We denote the final output of the network at each step as $O_{t}$ and its corresponding loss as $L_{O_t}$. Depending on the nature of the task, $L_{O_t}$ can be calculated as a reconstruction, segmentation or classification loss. Therefore, the resulting overall loss $L_{overall}$ is sum of contrastive loss, self-attention and contrastive predictions' losses at each step and downstream task loss, equation \ref{eqn:overall_loss}

\begin{equation}
\label{eqn:overall_loss}
L_{overall} = L_{F_c} + L_{C(t)} + L_{S(t)} + L_{O(t)}
\end{equation}

\vspace*{-0.15cm}
\section{Experiments}
\vspace*{-0.1cm}
\subsection{Dense Prediction}
\vspace*{-0.15cm}
In this section, we compare our method against several baselines for reconstruction and segmentation.  We evaluated our method on SUN360 \cite{song2015sun} and MS-COCO \cite{lin2014microsoft} datasets for reconstruction and ADE20k \cite{zhou2017scene} and COCO-Stuff \cite{caesar2018coco} for the segmentation task. We use Root-Mean-Squared-Error (RMSE) and mean pixelwise accuracy (mPA) to measure reconstruction and segmentation accuracy respectively and report the lowest RMSE and highest mPA for each experiment. 

Table \ref{tab:baseline} summarises our results on baselines that are trained using a variation of our method. These baselines help evaluation of each one of the modules in our network architecture independently. For all experiments in this table, we used 8 retina-like glimpses.

With No Contrast architecture, where we drop the contrastive loss, we observe a decrease in performance for reconstruction with a minimal decrement in performance on the segmentation task. This difference can be explained by how the contrastive loss is formulated in our architecture. For both reconstruction and segmentation tasks, the distance between the decoded representation from the layers in the contrastive stream and encoded representation from the corresponding layers in the ground truth stream in the U-Net architecture is minimized. This pushes the network to learn low-level features, which facilitates the reconstruction task. Whereas segmentation requires learning high-level features, to predict the class labels. Hence there is a minimal contribution of contrastive loss.

Using an additional full image segmentation decoder in the ground truth stream and minimizing the distance between its representation and that of corresponding layers in the contrastive stream may improve the performance. Although, this will make the ground truth stream twice as big. Hence to maintain the architectural consistency we keep this out of the scope of this work.

Random glimpse (SA) baseline keeps the network architecture intact while the selection of glimpses is made randomly instead of choosing the maximal value in the attention heatmap. This results in significant decrements in both reconstruction and segmentation performance compared to our model, proving the significance of attention-based glimpse selection. In the Random glimpse (No SA), the glimpse is selected randomly, and the self-attention module is replaced by a convolutional layer with an extra channel to maintain the total number of parameters in that layer. We observe that adding an extra-feature channel results in performance close to Random glimpse (SA) with a minor decrement in RMSE value on SUN360 and minor increment in that of MS-COCO. We also observe that segmentation performance improves for both datasets compared to Random glimpse (SA) close to that of our (full) model. The reason being more number of parameters in the next layer as a consequence of more number of input channels to that layer. Finally, we evaluate Random glimpse (No SAC), where we randomly select glimpses, use extra channel instead of attention maps, and drop the contrastive stream. We observe a decrease in performance compared to Random glimpse (No SA), with a minimal decrement in RMSE value for the MS-COCO dataset. Overall, we can conclude that the attention-based glimpse selection performs better than random glimpse selection, and contrastive stream results in better reconstruction performance.

\begin{table}[h]

\begin{center}
\begin{tabular}{|l|p{1cm}|p{3cm}|}
\hline
Dataset/Method& Ours & Attend and Segment \cite{seifi2020attend} \\
\hline\hline
Sun360 \cite{song2015sun} (RMSE) & \textbf{33.8} & 37.6 \\
\hline
MS-COCO \cite{lin2014microsoft} (RMSE) & \textbf{40.3} & 41.8 \\
\hline
ADE20k \cite{zhou2017scene} (mPA)& \textbf{0.524} & 0.479  \\
\hline
COCO-Stuff \cite{caesar2018coco} (mPA) & \textbf{0.478} & 0.456 \\
\hline
KITTI \cite{geiger2013vision} (mPA) & \textbf{0.806} & 0.805 \\
\hline
Cityscapes \cite{cordts2016cityscapes} (mPA) & 0.748 & \textbf{0.762} \\
\hline
CamVid \cite{brostow2008segmentation} (mPA) & 0.823 & \textbf{0.832} \\
\hline
\end{tabular}
\vspace{0.1cm}
\caption {\textbf{Active visual dense prediction}: Comparison of our model against Attend and Segment model \cite{seifi2020attend} on reconstruction and segmentation tasks, using root mean square error (RMSE) and mean pixel accuracy (mPA) evaluation metrics respectively.}
\label{tab:comparison}
\end{center}

\end{table}

\vspace*{-0.15cm}
\subsection{Comparison with Baseline}
\vspace*{-0.15cm}
Here we compare our model against the closest related work, the Attend and Segment model \cite{seifi2020attend}, for both reconstruction and segmentation tasks. It should be noted that the proposed model in \cite{seifi2020attend} is only evaluated for semantic segmentation, however for a wider evaluation, we train it on reconstruction task as well. Attend-and-Segment outperforms the previous reconstruction models (37.6 vs 39.0 \cite{ramakrishnan2018sidekick} and 38.8 \cite{jayaraman2018learning} RMSE) thus being the most competitive baseline to our work. For a fair comparison, we keep the number of parameters for both our model and Attend and Segment the same. From table \ref{tab:comparison}, it can be observed that our model achieves lower RMSE values than \cite{seifi2020attend} on both SUN360 and MS-COCO dataset.  

\begin{figure}
\vspace*{-0.7cm}
  \begin{center}
      \includegraphics[width=\linewidth]{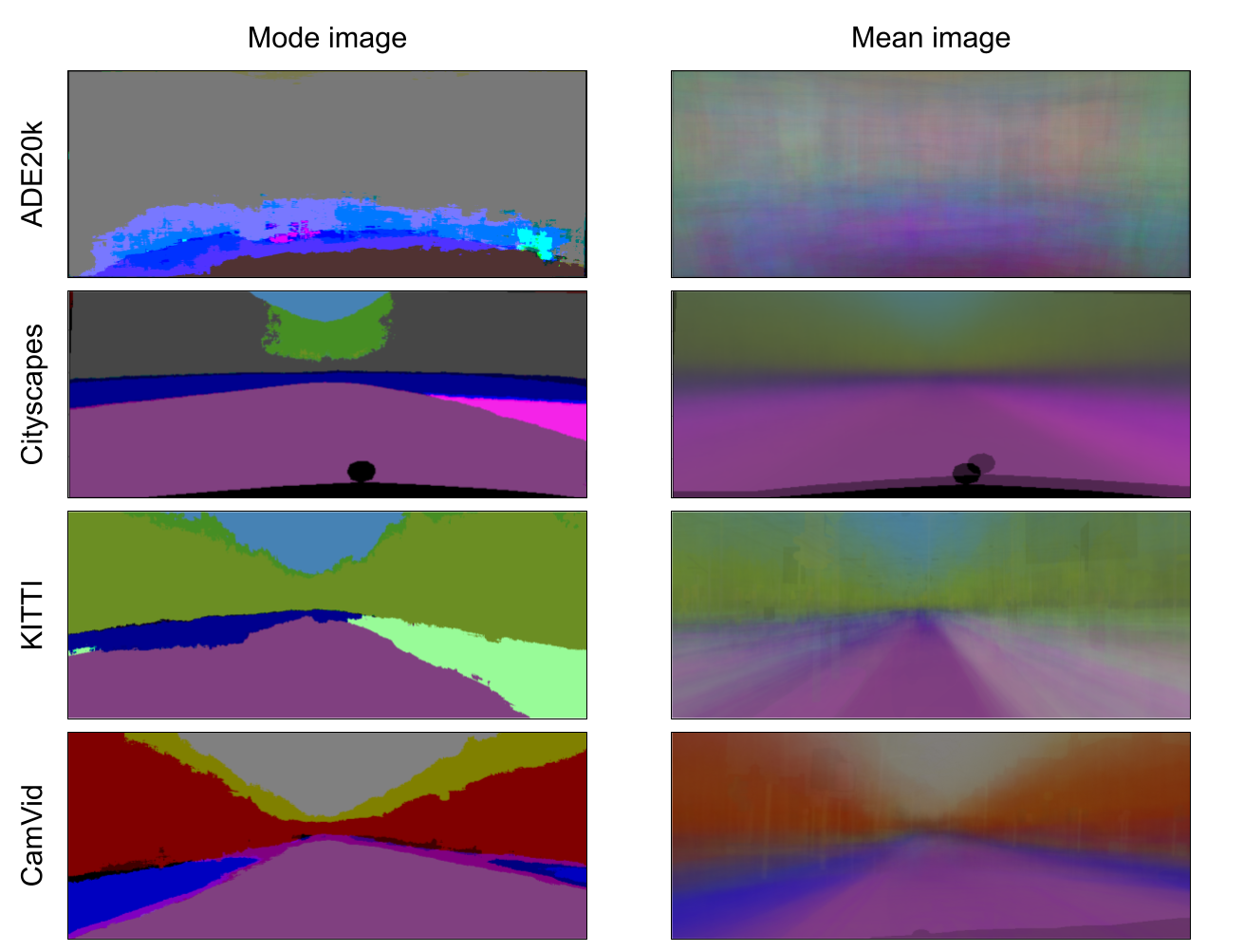}
   \end{center}
   \vspace*{-0.3cm}
      \caption{\textbf{Dataset bias:} (Left) represents the mode i.e. the most frequent labels and (right) represents the mean value of the ground truth labels. Mean and mode are computed for all the samples in the training set, except on Cityscapes dataset \cite{cordts2016cityscapes}, where a random set of 100 training samples are chosen. A mean image closer to mode represents homogeneity in the dataset, i.e., spatial bias in labels. ADE20k consisting of both indoor and outdoor scenes shows the least dataset bias. (More results on dataset bias can be found in supplementary).}
      \label{fig:mean_and_mode_image}
       \vspace*{-0.35cm}
   \end{figure}
For the segmentation task our model achieves a better mPA on ADE20k and COCO-Stuff while having similar performance on KITTI \cite{geiger2013vision}, Cityscapes \cite{cordts2016cityscapes} and CamVid \cite{brostow2008segmentation} datasets. 

\begin{figure}
  \begin{center}
      \includegraphics[width=\linewidth]{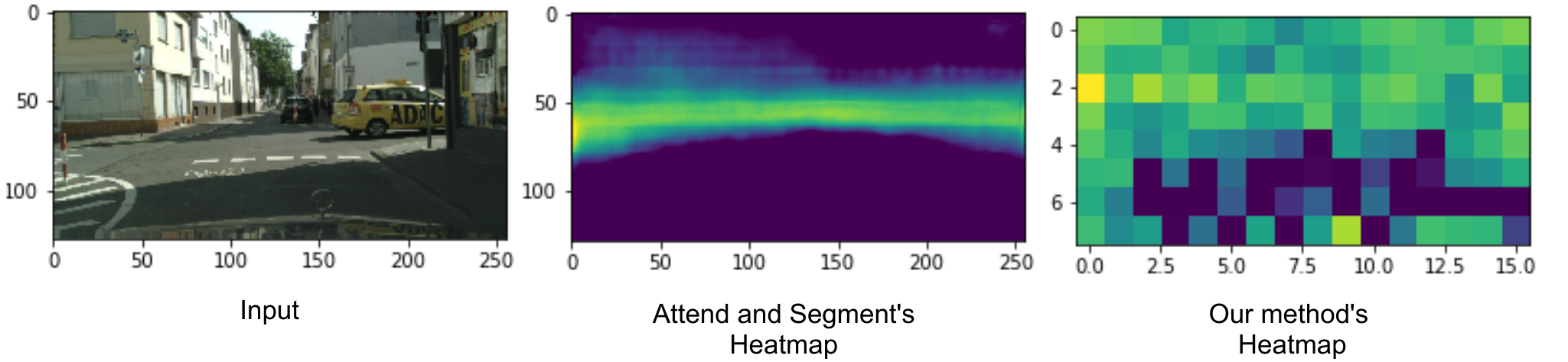}
   \end{center}
   \vspace*{-0.5cm}
      \caption{Comparison of the heatmaps generated by Attend and Segment and our method on Cityscapes dataset. Most images on this dataset consist of scenes with a road in front. Consequently, those areas are assigned less importance in the attention maps for both methods (i.e capturing dataset bias). Note that Attend and Segment's attention map looks smoother since it is generated in a higher resolution.}
      \label{fig:heatmapbias}
       \vspace*{-0.35cm}
\end{figure}

To analyze this further, we computed the mean and mode of the ground truth labels of the samples from these four datasets (figure \ref{fig:mean_and_mode_image}). We found that KITTI, Citiscapes, and CamVid show high homogeneity of the samples, with most of the images having a similar spatial arrangement of objects across the dataset. As shown by mode images in the figure \ref{fig:mean_and_mode_image}, most frequent labels are close to the mean of the images which implies less diversity between the samples (i.e high dataset bias). On the other hand, we find that the mode image of ADE20k is the least similar to its mean image, as ADE20k is a more diverse dataset consisting of both indoor and outdoor scenes. We also compute the attention map of both methods on randomly selected images from Cityscapes. Figure \ref{fig:heatmapbias} depicts that \cite{seifi2020attend} attention is largely concentrated around one region, while our self-attention module attends to a larger region in the image. Therefore, while our attention mechanism captures the bias in datasets with less diversity (darker areas in the figure) it performs better than \cite{seifi2020attend} on datasets like ADE20k which consists of various scenes and less dataset bias.

\begin{table}[h]
 \vspace*{-0.15cm}
\begin{center}
\begin{tabular}{|l|p{3.5cm}|}
\hline
Method & Classification Accuracy (\%) \\
\hline\hline
Ours (Full) & 56.4 \\
\hline
Ours (Full+Random) & 49.6 \\
\hline
Ours (No Decoder)& \textbf{67.2}   \\
\hline
Ours (No Decoder+Random) & 62.6 \\
\hline
Attend and Segment \cite{seifi2020attend} & 52.6 \\
\hline
\end{tabular}
\vspace{0.1cm}
\caption {\textbf{Active visual classification:} Comparison of classification performance on SUN360 dataset.}

\label{tab:classification}
\vspace*{-0.5cm}
\end{center}

\end{table}
 \vspace*{-0.25cm}
\subsection{Classification}
 \vspace*{-0.15cm}
To show the task adaptability, we evaluate our model on a 26-category classification task on SUN360 dataset, table \ref{tab:classification}. We compare our full model with Attend and Segment. To be consistent with the latter model, we set reconstruction as the auxiliary task, keep all the losses and add a separate classification head on the bottleneck features for both models. Under this framework, we observe an improvement of 3.8\% accuracy over Attend and Segment. In our (Full+Random) model, we use random glimpse selection instead of relying on the self-attention module. The accuracy decreases by 5.8\%, proving that the self-attention module learns to attend regions important for better classification.

Unlike Attend and Segment, our model does not rely on segmentation auxiliary task for glimpse selection. Hence we remove the reconstruction/segmentation decoder and only train the classifier, called the No Decoder model.
This results in a major improvement in accuracy of 10.8\% and 14.6\% over our (full) model and Attend and segment respectively. We repeat this experiment with random selection of glimpse. We observe a decrease in performance of 4.6\% over the No Decoder model, reconfirming the efficacy of our attention-based glimpse selection heuristic.

In particular our model without a decoder runs two times faster and uses a tenth of gpu memory compared to Attend and Segment (table \ref{timecomp}).

\begin{table}
\begin{center}
\begin{tabular}{|l||l|l|l|}
\hline
Method & Train(s) & Test(s)& GPU(GB) \\
\hline\hline
Attend and Segment & 474 & 45 & 5.5 \\
\hline
Ours (No Decoder)& \textbf{213}&\textbf{20}&0.4   \\
\hline
\end{tabular}
\vspace{0.1cm}
\caption{Comparison of Training/Test time and GPU memory usage (batch size: 6) for the classification task with Attend and Segment and our model without a decoder. Note that memory usage values are obtained with torch.cuda.max\_memory\_allocated() function.
}
\label{timecomp}
\end{center}
\vspace*{-0.5cm}
\end{table}


\begin{figure}
  \begin{center}
      \includegraphics[width=\linewidth]{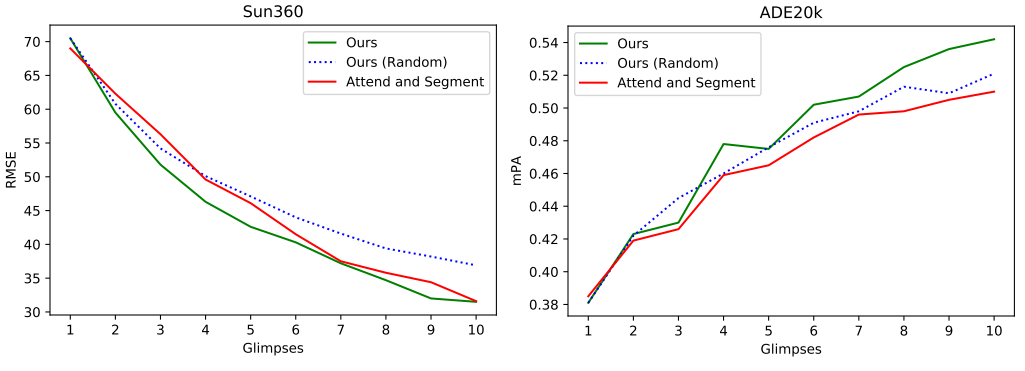}
   \end{center}
   \vspace*{-0.5cm}
      \caption{\textbf{Number of Glimpse vs performance:} (Left) shows RMSE values (y-axis) for the model trained and evaluated with different number of glimpse (x-axis) on SUN360 dataset, (right) shows the similar trend for mPA (y-axis) with different number glimpse (x-axis) on ADE20k dataset.}
      \label{ablation}
      \vspace*{-0.25cm}
\end{figure}

\begin{figure}
  \begin{center}
      \includegraphics[width=0.7\linewidth]{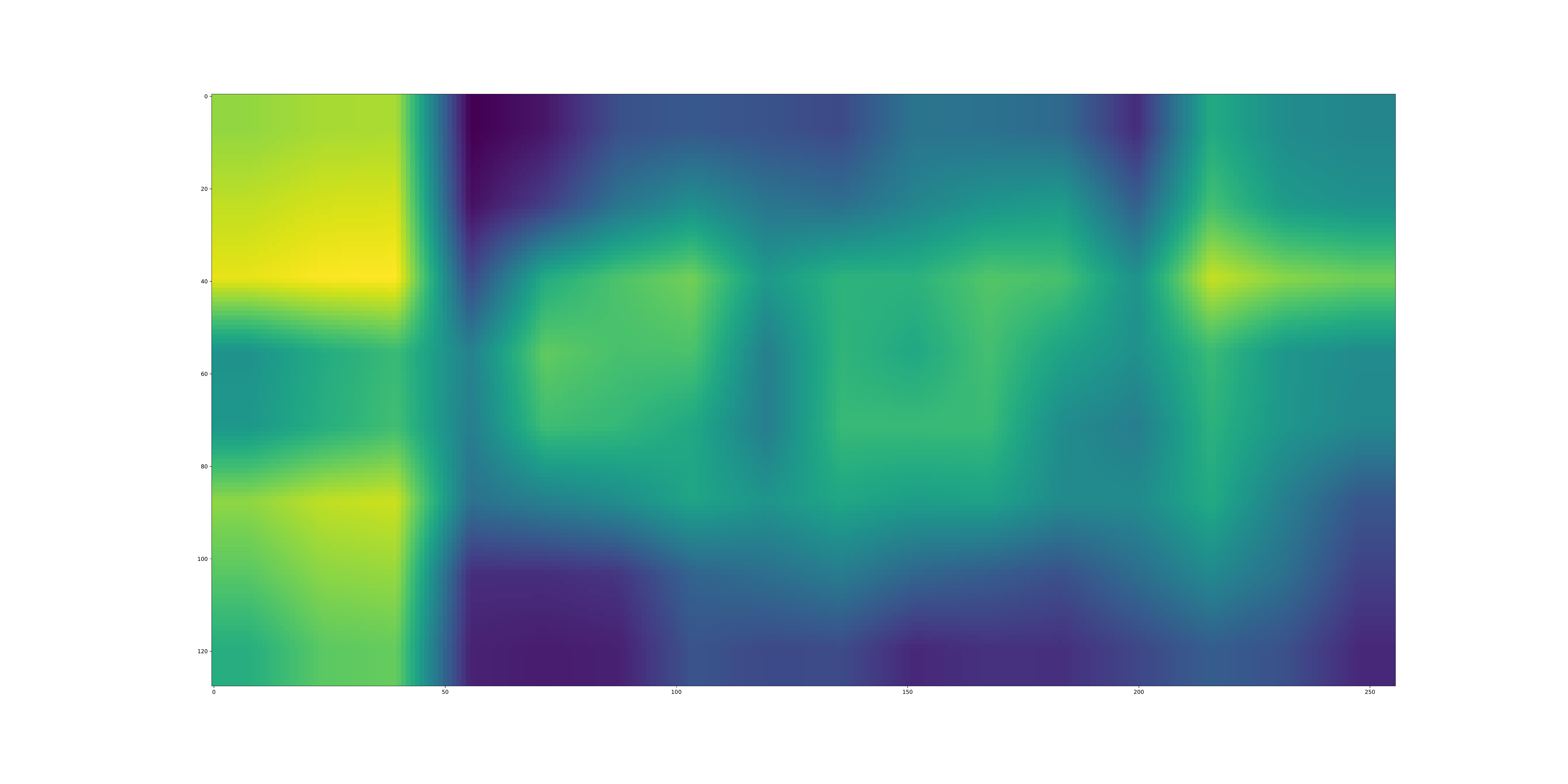}
   \end{center}
    \vspace*{-0.5cm}
    \caption{\textbf{Average glimpse image:} Average of the final glimpse-maps consisting of 8 glimpses over all the images of the SUN360 test set.
    }
      \label{fig:averageglimpse}
      \vspace*{-0.35cm}
\end{figure}

\begin{figure*}[h!]
  \begin{center}
      \includegraphics[width=\linewidth]{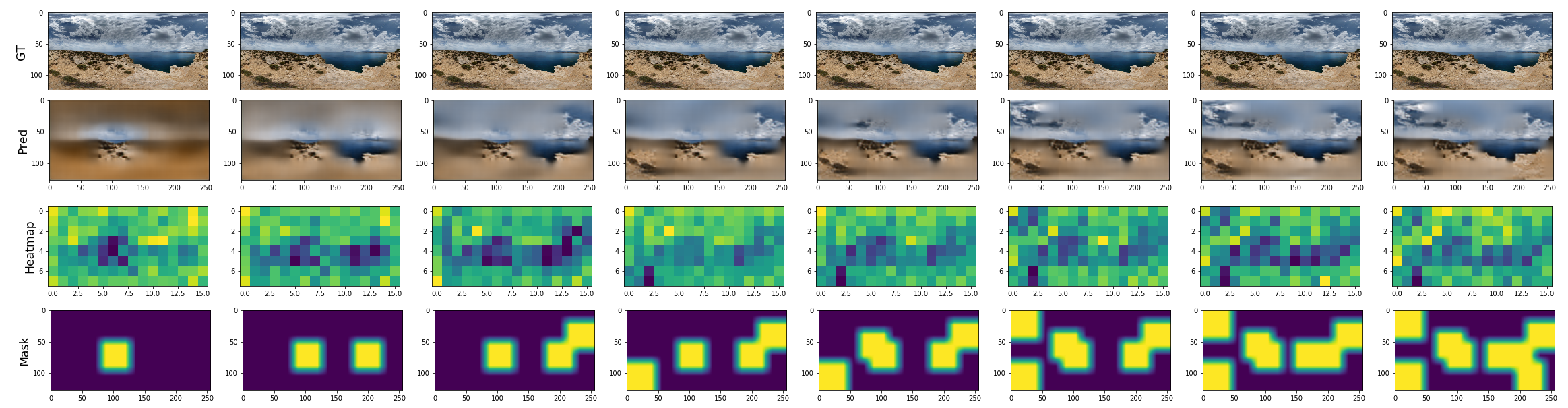}
   \end{center}
    \vspace*{-0.3cm}
    \caption{\textbf{Active visual reconstruction:} Step-by-step attended-glimpse based reconstruction of a natural scene (row 1) shows the ground truth image, (row 2) shows the reconstructed image at each step, (row 3) shows the attention heatmap of the last layer of the self-attention stream, (row 4) shows the input glimpse locations at each step.}
      \label{fig:steps}
      \vspace*{-0.5cm}
\end{figure*}

 \vspace*{-0.15cm}
\subsection{Glimpse Analysis}
 \vspace*{-0.15cm}
From the previous sections, we observe that our attention-based glimpse selection results in an overall improvement on both dense prediction and classification tasks. Hence, in this section, we provide a detailed analysis of the different properties of the glimpse.

\textbf{Number of glimpses:}
It is a important hyperparameter, decided primarily by the environment's difficulty and the agent's sensory capabilities inside the environment. Hence, studying the effect of the number of glimpses on the model's performance is important. While a large number of glimpses can observe a larger region of the environment, making the partial observability obsolete, too few glimpses do not provide enough information to reason about the environment. Figure \ref{ablation} compares three models, our model with attention-based glimpse selection, our model with random glimpse, and Attend and Segment model. For the reconstruction task on SUN360 dataset, our model's RMSE values are consistently lower when compared against the other two baselines. Poorest performance per number of glimpses is observed for random glimpse selection, suggesting that the self-attention module learns to look at uncertain regions in the image, and glimpse selection based on this plays a critical role in improving the performance.

For the segmentation task on ADE20k, the mPA is consistently higher with four or more glimpses for our model, representing a better segmentation performance against the baselines. For number of glimpses less than four, we find the performance of both attention-based and random glimpse selection to be better than Attend and Segment baseline.

\textbf{Average Glimpse Image:}
At each step, a glimpse is selected as the maximally activated region in the attention map. The underlying assumption being that region with maximum activation contributes most to the loss, and thereby attending to it decreases the loss and ambiguity in the reconstructed image. To investigate any pattern in the glimpse selection, we compute the average of the final glimpse-maps for all the images in the SUN360 test set trained for the reconstruction task. We observe that the top and bottom left corners are the most attended regions, shown by the brightest regions in the figure \ref{fig:averageglimpse}. As the SUN360 dataset consists of 360\textdegree view of either indoor or outdoor scenes, the network learns to look at these corners to reason about the general environment being indoor or outdoor. The model then uses the attended region to start painting sky or ground. It can also be seen from figure \ref{fig:steps}, from step 2 to 3, when the model attends the right corner in step 3, the ambiguity corresponding to all the sky regions from step 2 is cleared. Since the images are 360\textdegree views, the model learns to reason about right corner by looking at the left corner (more examples in supplementary material). Rest of the regions receives attention based on the image content shown by uniformly bright regions in the average glimpse image.

\begin{figure}[h!]
  \begin{center}
      \includegraphics[width=1.0\linewidth]{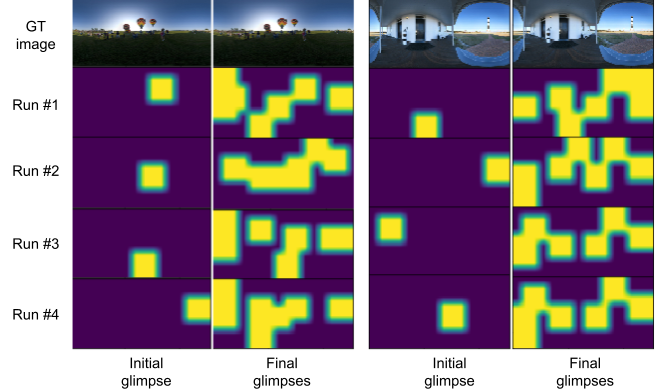}
   \end{center}
   \vspace*{-0.3cm}
      \caption{\textbf{Effect of glimpse initialization:} The first row shows the ground truth images randomly chosen from the SUN360 dataset, row 2-5 shows randomly chosen first glimpse location and the final set of attended glimpse location for four different runs.}
      \label{fig:init}
\end{figure}

\textbf{Effect of glimpse initialization:}
An image consists of specific spatial saliencies, and its reconstruction depends on attending those salient regions which are most ambiguous to reconstruct. Hence, a network that learns to attend for image reconstruction should attend those same regions irrespective of the first glimpse initialization. To verify this, we randomly initialize the first glimpse on images and observe the rest of the glimpses for those images. As shown in figure \ref{fig:init}, irrespective of the significant difference in spatial placement of the first glimpse across different runs, the final placement of glimpses shows high agreement. The high agreement of glimpses across the runs verifies that our model learns to attend to salient regions in the image.
\vspace*{-0.15cm}
 \vspace*{-0.15cm}
\subsection{Recurrent Glimpse Selection}
 \vspace*{-0.15cm}
As our final results, we show the step-by-step glimpse selection and reconstruction results of our method in figure \ref{fig:steps}. The glimpse mask added at each step in the bottom rows denote the visited location for that step. The attention map in the 3rd row
shows that the areas corresponding to the glimpse gets darker. These low attention values in that region are due to reduction in uncertainty after processing the glimpse. Based on the new information from the glimpse the decoder module improves the image reconstruction in each step. This iterative process of glimpse selection, reduces the uncertainty in each step to generate the final reconstructed image.
\vspace*{-0.25cm}
\section{Conclusion}
\vspace*{-0.15cm}
We proposed an attention-based active vision model that learns to attend the salient regions in the image based on the downstream task.
By disentangling the attention policy from the loss formulation and replacing it with our proposed self-attention module, we show that our architecture can be adapted to multiple tasks. 
We evaluate our model on reconstruction, segmentation, and classification on a diverse set of datasets and show a significant improvement in performance over the baseline model. Ablation study on individual modules of our model provided us with more insight about our performance gains. While contrastive stream resulted in a significant improvement for the reconstruction task, glimpse selection based on our self-attention module resulted in higher performance over randomly selected glimpses on all the tasks. The convergence of glimpse to similar image regions for different glimpse initialization suggests our model learns to attend to the salient regions in the scene. Lastly, we show how attending to the selected glimpses results in reduced uncertainty in the attention map, resulting in refinement of the image reconstruction at each step.

Finally, our results encourage further study on different aspect of this model. In particular, the memory module and the recurrent nature of the problem make it memory intensive. Reducing the memory needs of the architecture by means of representation learning is one future direction. Further study on contrastive loss for segmentation task is another interesting direction for the future research.
\vspace{-0.5cm}
\paragraph{\textbf{Acknowledgment}}
This work was supported by the FWO SBO project Omnidrone\footnote{https://www.omnidrone720.com/} and Flanders AI Research program\footnote{https://airesearchflanders.be/}.

{\small
\bibliographystyle{ieee_fullname}
\bibliography{egbib}
}

\end{document}